\pdfoutput=1

\documentclass[11pt]{article}

\usepackage{comment}
\newcommand{\transition}[1]{\texttt{TR(}{#1}\texttt{)}}
\newcommand{\reduce}[1]{\texttt{RE(}{#1}\texttt{)}}
\newcommand{\shift}{\texttt{SH}}
\newcommand{\boa}{\texttt{BOA}}
\newcommand{\eoa}{\texttt{EOA}}
\newcommand{\actionmem}{\text{SRU}}
\usepackage[shortlabels]{enumitem}
\usepackage[normalem]{ulem}
\useunder{\uline}{\ul}{}
\usepackage{multirow}
\usepackage{tabularx}
\usepackage{float}
\usepackage{dsfont}
\usepackage{mathtools}
\usepackage{booktabs}
\usepackage{multirow}
\usepackage{tikz, amsmath, amssymb}
\usepackage[labelfont=bf]{caption}
\usepackage{afterpage}
\usepackage{tikz}
\usepackage{xcolor}
\usetikzlibrary{fit, positioning, decorations.pathreplacing, calc}

\usepackage[linesnumbered, ruled, vlined]{algorithm2e}

\usepackage[final]{acl}

\usepackage{times}
\usepackage{latexsym}

\usepackage[T1]{fontenc}

\usepackage[utf8]{inputenc}

\usepackage{microtype}

\usepackage{inconsolata}

\usepackage{graphicx}

\title{Effective Multi-Task Learning for Biomedical Named Entity Recognition}

\author{João Ruano,\ Gonçalo M. Correia,\ Leonor Barreiros and Afonso Mendes \\
  Priberam Labs, Alameda D. Afonso Henriques, 41, 2º, 1000-123 Lisboa, Portugal \\
  \texttt{\{\href{mailto:joao.ruano@priberam.pt}{joao.ruano},\href{mailto:goncalo.correia@priberam.pt}{goncalo.correia},\href{mailto:leonor.barreiros@priberam.pt}{leonor.barreiros},\href{mailto:amm@priberam.pt}{amm}\}@priberam.pt}}

\begin{document}
\maketitle
\begin{abstract}
Biomedical Named Entity Recognition presents significant challenges due to the complexity of biomedical terminology and inconsistencies in annotation across datasets. This paper introduces SRU-NER (Slot-based Recurrent Unit NER), a novel approach designed to handle nested named entities while integrating multiple datasets through an effective multi-task learning strategy. SRU-NER mitigates annotation gaps by dynamically adjusting loss computation to avoid penalizing predictions of entity types absent in a given dataset.\footnote{Code is publicly available at \url{https://github.com/Priberam/sru-ner}.} Through extensive experiments, including a cross-corpus evaluation and human assessment of the model's predictions, SRU-NER achieves competitive performance in biomedical and general-domain NER tasks, while improving cross-domain generalization.

\end{abstract}

\section{Introduction}
Named entity recognition (\textbf{NER}) is a crucial step in several natural language processing pipelines, such as information extraction, information retrieval, machine translation, and question-answering systems~\cite{10.1007}. Given unstructured text, the task of NER is to identify and classify text spans according to categories of interest. These categories are defined depending on the downstream application and can range from general (\textit{people}, \textit{locations}, \textit{organizations}) to specific domains such as biomedical entities (\textit{genes}, \textit{diseases}, \textit{chemicals}).

In particular, Biomedical Named Entity Recognition (\textbf{BioNER}) is challenging due to the complexity of biomedical nomenclature. Morphologically, these entities can contain Greek letters, digits, punctuation (\textit{$\alpha$-tubulin}, \textit{IL-6}), form variations (\textit{inhibitor} vs. \textit{inhibitory}), and compound terms (\textit{tumor necrosis factor-alpha} vs. \textit{TNF-$\alpha$}). Semantically, polysemy (e.g., \textit{p53} referring to a gene, protein, or condition) adds ambiguity. These challenges make human annotation costly, leading to BioNER datasets that are smaller and often focus on a limited number of entity types~\cite{greenberg-etal-2018-marginal}.

One approach to addressing data scarcity while building a BioNER model is to leverage multiple datasets, each annotated with a specific subset of entities. However, simply training a single model on the union of all available datasets assumes that every entity type is consistently annotated across all training instances, which is not the case. This leads to a high prevalence of false negatives, as entities that are labeled in one dataset may be entirely ignored in another. On the other hand, training separate models for each dataset fails to exploit shared statistical patterns across datasets and introduces the challenge of resolving conflicting predictions at inference time \cite{greenberg-etal-2018-marginal}. Therefore, an effective strategy must balance learning from multiple sources while accounting for missing annotations and inconsistencies in labeling schemes.

Our contributions are three-fold: (i) we introduce \textbf{SRU-NER} (\textbf{S}lot-based \textbf{R}ecurrent \textbf{U}nit NER), a model which is able to solve nested NER through generating a sequence of actions; (ii) we propose an effective multi-task training strategy to handle the complex challenges of leveraging multiple NER datasets in a single model; and (iii) we show how the SRU-NER can handle multiple datasets on a single shared network through multiple experiments, including cross-corpus evaluations and a human evaluation on corpora of disjoint entity sets.

\begin{figure*}[t!]
    \centering
    \includegraphics[width=\linewidth]{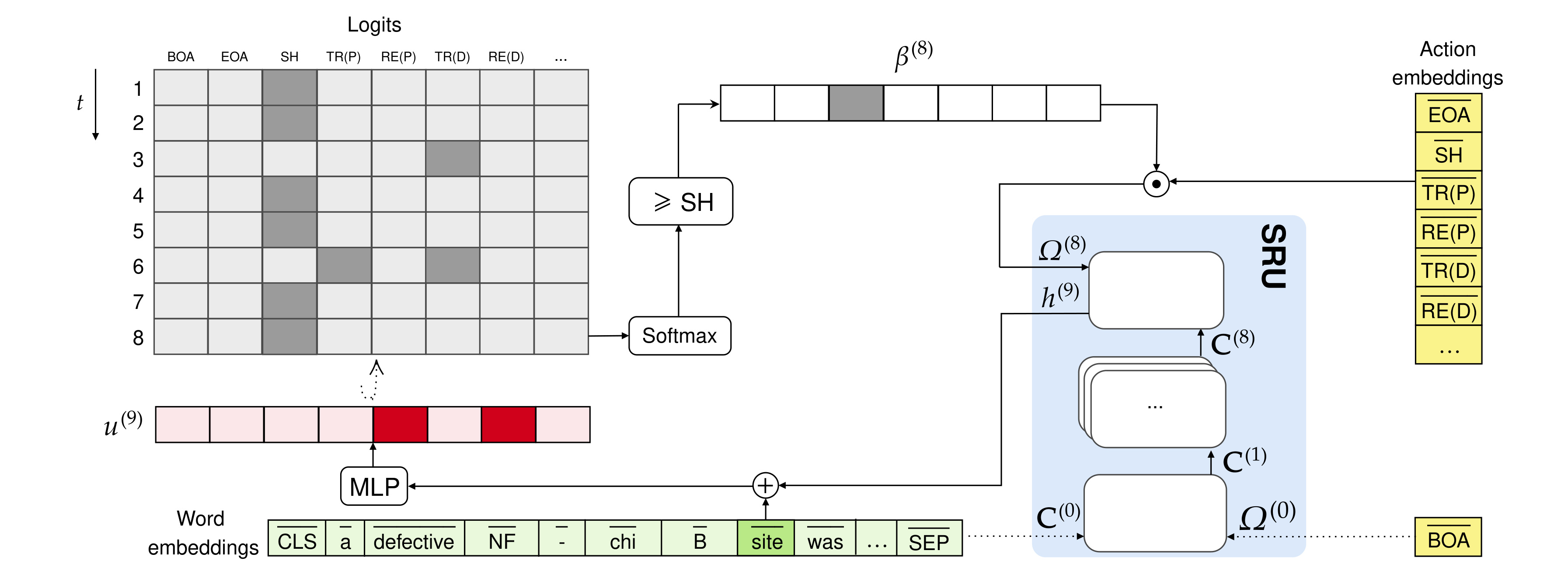}
    \caption{Action selection process for the sentence given in section \ref{sec:ac_enc}, at time step $t=9$. The gold nested mentions are "NF - chi B site", "chi B", of type \textit{DNA} (\textit{D}), and "NF - chi B" of type \textit{Protein} (\textit{P}). To compute the logits $u^{(9)}$, the model leverages the logits of the previous time steps, action embeddings and word embeddings.}
    \label{fig:model}
\end{figure*}

\section{Related work}

Named entity recognition has evolved significantly in the last decades. Early systems relied on rule-based methods, which were interpretable but lacked flexibility. The introduction of machine learning enabled more adaptable approaches, further enhanced by deep learning techniques that captured complex linguistic patterns. Recently, Transformer-based architectures have set new benchmarks, driving significant advancements in NER performance~\cite{survey_li,keraghel2024recentadvancesnamedentity}. In the CoNLL-2003 dataset~\cite{conll_tjong-kim-sang-de-meulder-2003-introduction}, a benchmark for NER tasks, performance has improved substantially, with F1 scores that have soared above 94\%~\cite{ACE_wang-etal-2021-automated}. The same phenomenon is seen for the GENIA corpus~\cite{genia_10.1093/bioinformatics/btg1023}, a nested BioNER dataset, with test F1 scores exceeding 80\%~\cite{yu-etal-2020-named,ijcai2021p0542, shen-etal-2021-locate, PIQN_shen-etal-2022-parallel}.

To tackle the proliferation of BioNER datasets, several studies have turned to multi-task learning (\textbf{MTL}; \citealp{review_flat_nested}). Traditional deep learning NER models trained on a single dataset are referred to as \textit{single-task} models, as they specialize in identifying mention spans for the specific entity types annotated within their training data. Single-task models often underperform on out-of-domain settings. In contrast, MTL frameworks leverage multiple datasets, each corresponding to a different \textit{task}, allowing the model to learn from diverse sources. The fundamental premise is that different datasets share information which can be jointly leveraged to encourage the learning of more generalized representations, hence improving a model's robustness~\cite{10.1007/978-3-030-35166-3_31, survey_li}.

MTL learning frameworks can be categorized into two types, depending on which modules are shared across tasks: (i) those that share the encoding layers while maintaining task-specific decoding layers~\citep{crichton_neural_2017, cross_type_wang, khan2020mtbionermultitasklearningbiomedical}, and (ii) those that share \textit{all} layers~\citep{greenberg-etal-2018-marginal, huang-etal-2019-learning, banerjee_biomedical_2021, aioner, taughtnet}. SRU-NER resembles models of type (ii), which share its decoding layers across all tasks. Typically, these models have a natural problem with false negatives, as the unified decoder may struggle to distinguish task-specific entity boundaries and labels, leading to the omission of valid entities. Our approach avoids this issue through an effective multi-task learning strategy.

\section{Effective Multi-Task Learning for Named Entity Recognition}

The proposed model, SRU-NER, solves the task of nested named entity recognition similar to that of a transition-based parser \cite{dyer-etal-2015-transition, marinho-etal-2019-hierarchical}.
Given a sequence of words ${S=[w_1, w_2, \ldots, w_N]}$, the model generates a sequence of \textit{actions}. At each time step, the actions are chosen depending on the words of the sentence and on the previously chosen actions. At the end of the parsing procedure, the complete sequence of actions is decoded into mentions.

\subsection{Action encoding}
\label{sec:ac_enc}

Consider the system is trained to recognize mentions of entity types belonging to ${\mathbb{E}=\{e_1, e_2, \ldots, e_M\}}$.  Let $\mathcal{A}_{\mathbb{E}}$ stand for the system's $2M+2$ possible \textit{actions}: two special tokens ($\shift$ and $\eoa$) and, for each entity type $e_i$, a pair of actions denoted $\transition{e_i}$ and $\reduce{e_i}$. $\transition{e_i}$, short for "transition to entity $e_i$", indicates the start of a mention of type $e_i$; one says that this action \textit{opened} a mention of type $e_i$. $\reduce{e_i}$, short for "reduce of entity $e_i$", indicates the end of the mention of type $e_i$ that was opened more recently; one says that a mention was \textit{closed} by this action. $\shift$, short for "shift", indicates that the input pointer should move to the next token; therefore, there is one $\shift$ for each word in the sentence. Finally, $\eoa$ is the end action.

These actions encode nested mentions effectively through the order in which they are chosen.
If a mention of type $e_k$ starts at the word $w_i$ and ends at the word $w_j$, $\transition{e_k}$ appears before the $\shift$ representing the $i$-th word, and $\reduce{e_j}$ appears after the $\shift$ representing the $k$-th word;
if two mentions start at the same word, the $\transition{}$ of the longest mention appears first; conversely, if two mentions end at the same word, the $\reduce{}$ of the shortest mention appears first. Consider the following sentence from the GENIA dataset~\cite{genia_10.1093/bioinformatics/btg1023}:

\vspace{1.5cm}

\begin{tikzpicture}[remember picture, overlay]

\node[text width=12cm, anchor=west] at (0,0) {
    \textit{a} \textit{defective} \textit{NF} \textit{-} \textit{chi} \textit{B} \textit{site} \textit{was} \textit{completely} $\ldots$
};

\draw[blue, ultra thick] (1.95,0.6) -- (3.7,0.6); 
\node[blue] at (2.5,1.1) {Protein};

\draw[orange, ultra thick] (2.8,0.4) -- (3.7,0.4); 

\draw[orange, ultra thick] (1.95,0.8) -- (4.35,0.8); 
\node[orange] at (4,1.1) {DNA};

\end{tikzpicture}

\vspace{0.5cm}

\noindent This sentence has nested mentions, \textit{e.g.} the mention "NF - chi B" of type \textit{Protein} is contained in the mention "NF - chi B site" of type \textit{DNA}. The action encoding of the sentence with its mentions is:
$\shift$ $\to$ $\shift$ $\to$ $\transition{\textit{DNA}}$ $\to$ $\transition{\textit{Protein}}$ $\to$ $\shift$ $\to$ $\shift$ $\to$ $\transition{\textit{DNA}}$ $\to$ $\shift$ $\to$ $\shift$ $\to$ $\reduce{\textit{DNA}}$ $\to$ $\reduce{\textit{Protein}}$ $\to$ $\shift$ $\to$ $\reduce{\textit{DNA}}$ $\to$ $\shift$ $\to$ $\shift$ $\to$ $\ldots$ $\to$ $\eoa$.

\subsection{Overall architecture}
\label{sec:overall_arch}
Using the previous notation, suppose one wants to detect mentions of $\mathbb{E}$ in the sentence $S$.
The model consists of three consecutive steps: the encoding of $S$ into a dense contextual embedding matrix $\mathbf{S}$, the iterative action generation procedure, and the decoding of the chosen actions into the mentions present in the sentence.

\paragraph{Contextual embeddings} For the first step, $S$ is passed through a BERT-like encoder to generate a matrix of contextual embeddings. For each word $w_i$, its dense embedding, denoted by $\overline{w_i}$, is obtained by max-pooling across the embeddings of its subwords. In this way, the encoded sentence $\mathbf{S}$ is a tensor of size $(N+2, d_{\text{enc}})$, ${\mathbf{S}=\left[\overline{\text{CLS}}, \overline{w_1}, \overline{w_2}, \ldots, \overline{w_N}, \overline{\text{SEP}}\right]}$,
where $d_{\text{enc}}$ is the encoder embedding dimension, $\overline{\text{CLS}}$ (respectively $\overline{\text{SEP}}$) is the embedding of the classification (respectively, separator) token of the encoder.

\paragraph{Action generation} Given $\mathbf{S}$, the model enters an iterative action selection process, where at each time step $t$, logits are computed for each possible action in $\mathcal{A}_{\mathbb{E}}$.\footnote{Unlike token-based labeling approaches, the total number of time steps is not determined \textit{a priori}, although always bounded below by $N$, the number of words in $S$.} Figure~\ref{fig:model} shows a schematic representation of a time step of the cycle.

\par More concretely, define $u_{a_i}^{(t)}$ to be the logit value of action $a_i\in \mathcal{A}_{\mathbb{E}}$ for time step $t$. Suppose the system has already computed these values for the first $T\geq 1$ time steps, and is therefore about to compute them for time step $t=T+1$. According to the last section, the $\shift$ action corresponds to advancing a token in the sentence $S$. Hence, define

\begin{equation}
    p^{(t)} = \sum_{t_0\,\leq\, t} \mathds{1}\left(\arg\max\limits_{a_i\in \mathcal{A}_{\mathbb{E}}} \left(u_{a_i}^{(t_0)}\right) = \shift\right),
    \label{eq:word_pointer}
\end{equation}

\noindent where $\mathds{1}$ stands for the indicator function. $p^{(t)}$ is therefore the number of tokens that have already been parsed at a previous time step $t$, for ${1 \leq t\leq T}$. Lastly, define, for each ${1 \leq t\leq T}$,

\begin{equation}
    \Omega^{(t)} = \sum\limits_{a_i\in \mathcal{A}_{\mathbb{E}}}\beta_{a_i}^{(t)} \ \, \overline{a_i},
    \label{eq:action_embed}
\end{equation}

\noindent where $\overline{a_i}$ is a trained embedding of size $d_{\text{enc}}$ and

\begin{align*}
        \beta_{a_i}^{(t)} = \begin{cases}
        u_{a_i}^{(t)} & \text{if}\ u_{a_i}^{(t)} \geq u_{\shift}^{(t)} \\
        0 & \text{otherwise}
        \end{cases}
\end{align*}

\noindent In other words, $\Omega^{(t)}$ is a weighted embedding of the actions chosen at time step $t$, where actions with logits lower than the logit of $\shift$ are excluded.

Let $\mathbf{u}^{(T+1)}$ be the vector of logits $u_{a_i}^{(T+1)}$ over $a_i \in \mathcal{A}_{\mathbb{E}}$. These are computed as

\begin{equation}
    \mathbf{u}^{(T+1)} = \text{MLP}\left(f\left(p^{(T)},\Omega^{(T)}\right)\right),
\label{eq:logits}
\end{equation}

\noindent where the $\text{MLP}$ is composed of a dropout layer, a fully-connected layer, a $\tanh$ activation, and a linear layer with output nodes corresponding to each action in $\mathcal{A}_{\mathbb{E}}$. The input of this MLP is

\begin{equation*}
    f\left(p^{(T)},\Omega^{(T)}\right) = \mathbf{S}_{p^{(T)}+1} \oplus \actionmem\left(\Omega^{(T)}\,,\, p^{(T)}\right)\, ,
\end{equation*}

\noindent \textit{i.e.} the concatenation of the embedding of the \textit{next} token, $\mathbf{S}_{p^{(T)}+1}$, and an embedding of the last state of a "processed actions memory". This memory holds an action history and computes weighted embeddings at each call by leveraging a set of internal latent representations. This module is refered to as the \textbf{S}lot-based \textbf{R}ecurrent \textbf{U}nit (\textbf{SRU}), and is described in section~\ref{sec:SRU}.

In order to make the first prediction, $\mathbf{u}^{(1)}$, the system is initialized by setting $p^{(0)} = 0$, and $\Omega^{(0)}$ to be another trained embedding of size $d_{\text{enc}}$, denoted by $\overline{\boa}$.\footnote{In this text, a zero-indexing notation is adopted for tensors, and so $\mathbf{S}_{p_{0} + 1} = \overline{w_1}$.} The action generation cycle terminates when a time step $t=T_{\text{final}}$ is reached such that

\begin{equation}
    \label{eq:action_generation_terminates}
    \operatorname{Sigmoid}\left(u_{\eoa}^{(T_{\text{final}})}\right) > 0.5\ .
\end{equation}

\paragraph{Decoding} At the end of the action generation cycle, the output logits from all time steps are passed through a sigmoid function. This produces a set of independent probability scores for each action in $\mathcal{A}_{\mathbb{E}}$, from which mention spans are extracted. The decoder module maintains separate stacks of open spans for each entity type in $\mathbb{E}$, allowing spans of different types to overlap.  

The decoding process iterates through the list of probability scores until reaching a time step where the highest-scoring action is $\eoa$\footnote{This stopping condition was shown to provide better results empirically, despite being different to that of the action generation procedure, present in equation (\ref{eq:action_generation_terminates}).}. Before such a time step is reached, the decoder proceeds following two rules: (i) if the highest-scoring action is $\shift$, a pointer that counts the number of parsed words is incremented; and (ii) if the highest-scoring action is a $\transition{}$ or a $\reduce{}$, the entity mention stacks are updated.
In the latter case, only actions with probability scores above $0.5$ are considered. Transition actions open new spans, while reduce actions close the most recent span of the corresponding entity type, as discussed in section \ref{sec:ac_enc}.  

\begin{figure}[t!]
    \centering
    \includegraphics[width=1\linewidth]{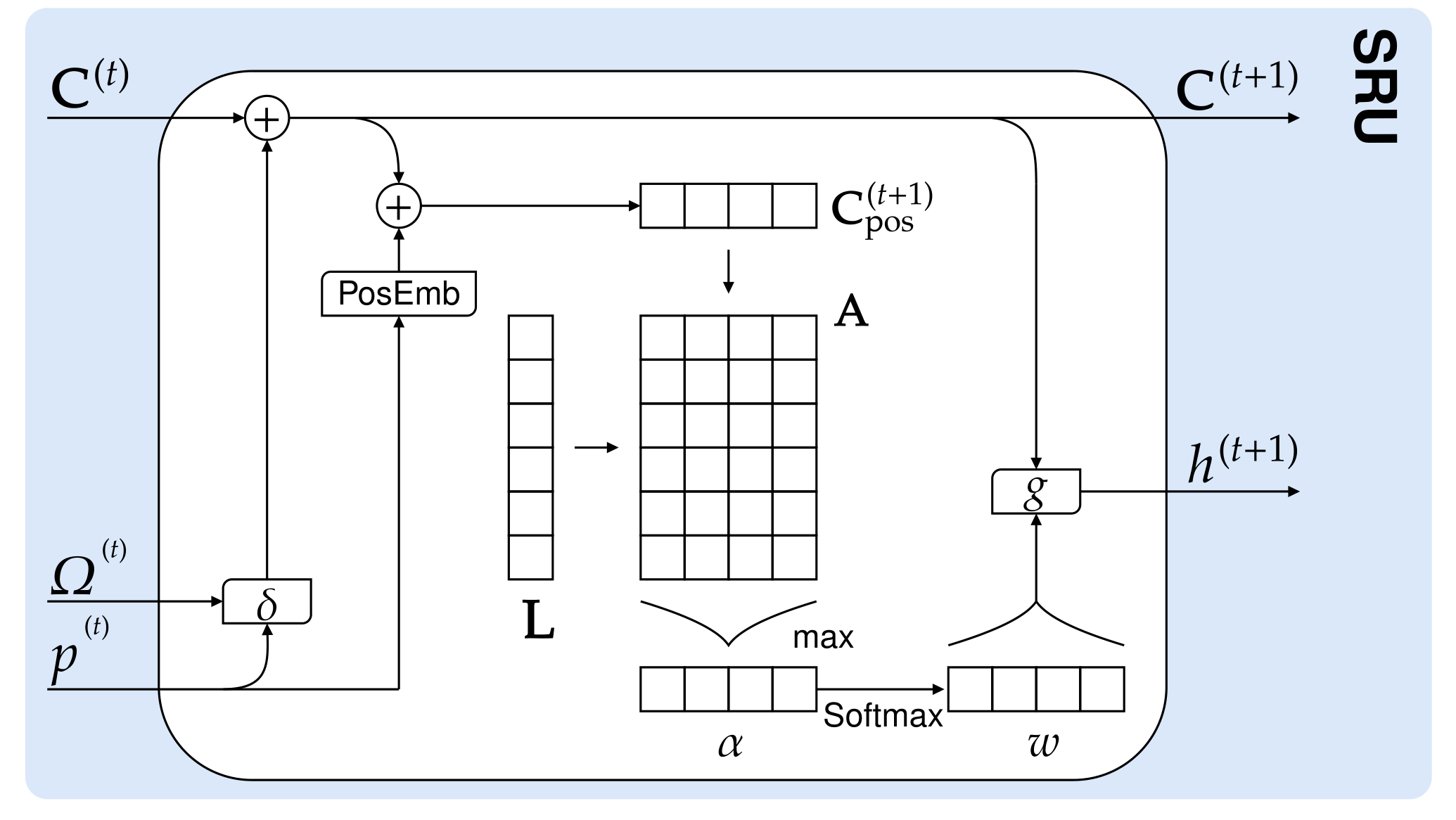}
    \caption{SRU unit at time step $t$. Its internal state is updated depending on its current state $\mathbf{C}^{(t)}$ and the weighted action embeddings $\Omega^{(t)}$. This stateful function also leverages a set of latent representations. It produces an output embedding $h^{(t+1)}$ by applying an attention mechanism to the updated state.}
    \label{fig:SRUcell}
\end{figure}

\subsection{Slot-based Recurrent Unit}
\label{sec:SRU}

The Slot-based Recurrent Unit ($\actionmem$) is a stateful function that, at each time step, takes a pair of inputs, updates its internal state, and produces an output embedding.


At each time step $t$, the $\actionmem$ updates its internal state according to

\begin{equation*}
    \mathbf{C}^{(t+1)} = m\left(\mathbf{C}^{(t)},\,\Omega^{(t)},\, p^{(t)}\right)\, ,
\end{equation*}

\noindent where $\mathbf{C}^{(t)}\in \mathbb{R}^{Q\times d}$ is the SRU's internal state matrix, $\Omega^{(t)}\in \mathbb{R}^d$ is an input vector, and ${p^{(t)}\in \{0,1,\ldots, Q-1\}}$ is an input integer. It also produces an output embedding $h^{(t+1)}\in \mathbb{R}^d$ via

\begin{equation*}
    h^{(t+1)} = g\left(\mathbf{C}^{(t+1)},\, p^{(t)}\right)\, .
\end{equation*}

\noindent A schematic representation is present in Figure \ref{fig:SRUcell}. $Q$ and $d$ refer to the number of rows (or \textit{slots}) in the internal state matrix and the hidden dimension of the input and output embeddings, respectively.


\par The function $m$ updates $\mathbf{C}^{(t)}$ by summing the input vector $\Omega^{(t)}$ to its $p^{(t)}$-th row, \textit{i.e.}

\begin{equation*}
m\left(\mathbf{C}^{(t)},\,\Omega^{(t)},\, p^{(t)}\right) \coloneqq \mathbf{C}^{(t)} + \delta_{p^{(t)}}\, \left(\Omega^{(t)}\right)^{T}
\end{equation*}

\noindent where $\delta_{p^{(t)}} \in \mathbb{R}^Q$ is a one-hot vector with $1$ in its $p^{(t)}$-th coordinate.

\par The output embedding $h^{(t)}\in \mathbb{R}^d$ is obtained via the function $g$, defined as

\begin{equation*}
    g\left(\mathbf{C}^{(t+1)},\, p^{(t)}\right) \coloneqq \mathbf{w}^{T}\, \left(\mathbf{C}^{(t+1)} \, \mathbf{D}_1\right)
\end{equation*}

\noindent where $\mathbf{D}_1$ is a trained diagonal matrix of size $d$ and $\mathbf{w}\in \mathbb{R}^Q$ are weights computed via an attention mechanism inspired by \citealp{ganea-hofmann-2017-deep}, detailed as follows. First, $\mathbf{C}^{(t+1)}$ is enhanced by adding positional information,

\begin{equation}
    \label{eq:sum_pos_embeds}
    \mathbf{C}^{(t+1)}_{\text{pos}} = \alpha\ \mathbf{C}^{(t+1)} + \operatorname*{Dropout}\left(\mathbf{P}\left(p^{(t)}\right)\right)
\end{equation}

\noindent where $\alpha$ is a trained scaling parameter, and $\mathbf{P}\left({p^{(t)}}\right)\in \mathbb{R}^{Q\times d}$ are positional embeddings.\footnote{These positional embeddings are \textit{relative}, in the sense that each row of $\mathbf{P}\left({p^{(t)}}\right)$ is selected from a table of trained embeddings based on its distance to the row with index $p^{(t)}$.} Next, a set of $J$ trained latent embeddings of size $d$ are used to compute an attention score for each row in $\mathbf{C}^{(t+1)}$. Defining $\mathbf{L}\in \mathbb{R}^{J\times d}$ to be the matrix of latent embeddings, an attention score matrix is computed by

\begin{equation*}
    \mathbf{A} = \operatorname*{Dropout}(\mathbf{L})\ \mathbf{D}_2\ \left(\mathbf{C}^{(t+1)}_{\text{pos}}\right)^T,
\end{equation*}

\noindent where $\mathbf{D}_2$ is a trained diagonal matrix of size $d$. An attention score for each slot is obtained by setting $\mathbf{s} = \max\limits_{j}(A_{jq})$ for ${q\in \{0,1,\ldots,Q-1\}}$. Finally, the scores $\mathbf{s}$ are normalized through a softmax to get the weights $w\in \mathbb{R}^{Q}$.

The $\actionmem$ module is used at each action generation time step to compute an embedding that models the current state of a "processed actions memory" stack. For each time step $t$, the input integer $p^{(t)}$ is the one defined by equation (\ref{eq:word_pointer}), and the input vector $\Omega^{(t)}$ is the one defined by equation (\ref{eq:action_embed}). Furthermore, $d$ is set to be the encoder embedding dimension $d_{\text{enc}}$, the number of slots to be $Q = N+2$, and the number of latent variables $J$ to be an integer multiple\footnote{For the experiments conducted, it was set to $2$ or $10$ (see Table \ref{tab:hyper} in Appendix \ref{appendix:training}).} of $|\mathcal{A}_{\mathbb{E}}| = 2M+2$. The internal state matrix is initialized by setting $\mathbf{C}^{(0)}=\mathbf{S}$. Taking this choice of initialization into account, and referring back to equation (\ref{eq:logits}), for the computation of $h^{(T+1)}= \actionmem\left(\Omega^{(T)}\,,\, p^{(T)}\right)$, all the slots of the updated internal state matrix $\mathbf{C}^{(T+1)}$ are filled with the embeddings of the encoded sentence $\mathbf{S}$. In addition, a history of the previously chosen actions is present in $\mathbf{C}^{(T+1)}$ since, at each call of the $\actionmem$ module in previous time steps $0\leq t \leq T$, the weighted action embeddings $\Omega^{(t)}$ of equation (\ref{eq:action_embed}) were summed to the slots pointed to by $p^{(t)}$.

\section{Multi-task training strategy}
\label{sec:training_strategy}
\par Suppose the model is trained on an ensemble of $K$ datasets $\mathcal{D}=\{D_i\}_{i=1}^{K}$, where each dataset $D_i$ is annotated with spans of entity types $\mathbb{E}_i$. In order to account for differences in labeling schemes, during training, the entity types of distinct datasets are always considered to be distinct as well.\footnote{In practice, this is implemented by simply changing the name of an entity type $e\in \mathbb{E}_i$ belonging to $D_i$, to $i\_e$ in $\mathbb{E}$.} Therefore, the model is trained to recognize spans of entity types in the disjoint union set $\widehat{\mathbb{E}}=\bigsqcup_{i=1}^K \mathbb{E}_i$.

\begin{table*}[t!]
\small
\centering
    \begin{tabular}{lrrrrrr}
    \toprule
\multirow{2}{*}{Dataset} & \multicolumn{2}{c}{SRU-NER} & \multicolumn{1}{c}{\multirow{2}{*}{\citealp{cross_type_wang}}} & \multicolumn{1}{c}{\multirow{2}{*}{\citealp{huang-etal-2019-learning}}} & \multicolumn{1}{c}{\multirow{2}{*}{\citealp{khan2020mtbionermultitasklearningbiomedical}}} & \multicolumn{1}{c}{\multirow{2}{*}{\citealp{taughtnet}}} \\
                         & \multicolumn{1}{c}{\textit{Merged}}     & \multicolumn{1}{c}{\textit{Disjoint}}     & \multicolumn{1}{c}{}                               & \multicolumn{1}{c}{}                              & \multicolumn{1}{c}{}                               & \multicolumn{1}{c}{}                                 \\ \midrule
BC2GM                    & 78.80                               & {\ul 83.95}                           & 80.74 *                                                                       & 79.1                                                                                 & 83.01 *                                                                                                 & \textbf{84.84}                                                        \\
BC4CHEMD                 & {\ul90.42}                     & \textbf{92.05}                           & 89.37 *                                                                       & 87.3                                                                                 & ---                                                                                                     & ---                                                                   \\
BC5CDR                   & 89.37                         & \textbf{90.26}                        & 88.78 *                                                                       & ---                                                                                  & {\ul 89.50} *                                                                                                 & $\diamond$                                                            \\
JNLPBA                   & 72.15                               & {\ul 76.00}                           & 73.52 *                                                                       & \textbf{83.8}                                                                        & 72.89 *                                                                                                 & ---                                                                   \\
Linnaeus                 & \multicolumn{2}{c}{\textbf{88.82}}                                          & ---                                                                           & {\ul 83.9}                                                                           & ---                                                                                                     & ---                                                                   \\
NCBI Disease             & 87.32                               & {\ul 88.71}                           & 86.14 *                                                                       & 84.0                                                                                 & 88.10 *                                                                                                 & \textbf{89.20}                                                        \\ \midrule
\textit{Average}         & 84.48                               & 86.63                                 &                                                                               &                                                                                      &                                                                                                         &                                                                       \\ \bottomrule                                            
\end{tabular}
\caption{Micro-F1 scores of several multi-task models trained on subsets of an ensemble of six biomedical datasets. For SRU-NER, scores are reported by considering two evaluation scenarios, \textit{Merged} and \textit{Disjoint}, as explained in section \ref{sec:results_MT}. Best scores are \textbf{bold}, and second best scores are \underline{underlined}. Symbol reference:\\
--- : dataset was absent in training;\\
* : model was trained on both the training and development splits of the corpora;\\
$\diamond$ : model was trained using only the 'Chemical' annotations of BC5CDR, obtaining an F1 of 93.95; for the same tag, SRU-NER gets an F1 of 93.77 on the disjoint evaluation and 93.18 on the merged evaluation.
}
\label{tab:MT_results}
\end{table*}

The training objective of the model is to minimize the mean loss of the samples in a batch. Each batch is constructed by randomly selecting samples from $\mathcal{D}$. To ensure a balanced contribution from all datasets, the probability of selecting a sample from a given dataset is inversely proportional to the total number of sentences in that dataset. The total number of samples per epoch is the average number of sentences in the datasets of $\mathcal{D}$.

Let $S$ be a sentence in the batch, coming from dataset $D_i$, and thus annotated with gold spans of entity types $\mathbb{E}_i$. The output of the action generation cycle is a matrix

\begin{equation*}
\mathbf{U}=\left(u_{a_i}^{(t)}\right)_{t=1,\, \ldots\, ,\, T_{\text{final}}\, ;\, a_i\in \mathcal{A}_{\mathbb{E}}},
\end{equation*}

\noindent where each row $u^{(t)}_{\ast}$ contains the logits, for time step $t$, over \textit{all} actions $\mathcal{A}_{\widehat{\mathbb{E}}}$ associated with the disjoint union set $\widehat{\mathbb{E}}$.\footnote{At inference time, the action generation procedure halts when the probability of the $\eoa$ action exceeds a threshold, as described in section \ref{sec:overall_arch}. However, during training, in order to guarantee that all gold actions are considered, the cycle halts only after all tokens have been parsed (\textit{i.e.} shifted).} To compute a loss value for $\mathbf{U}$, the following constraints are enforced:
\begin{enumerate}[i)]
    \item on one hand, the model \textit{should} be penalized for failing to predict the $\transition{}$ and $\reduce{}$ actions that correspond to the gold spans of the entity types $\mathbb{E}_i$, for which $S$ is annotated; but
    \item on the other hand, the model \textit{should not} be penalized for predicting $\transition{}$ and $\reduce{}$ actions of entity types in $\widehat{\mathbb{E}}\setminus \mathbb{E}_i$, which are not annotated in $S$.
\end{enumerate}

 \par In practice, this strategy is applied as follows. The list of actions corresponding to the gold annotations of sentence $S$ (constructed as detailed in section \ref{sec:ac_enc} and considering the disjoint entity type set $\widehat{\mathbb{E}}$) is augmented to a matrix ${\mathbf{G}=\left(G_{a_i}^{(t)}\right)\in \mathbb{R}^{T_{\text{initial}}\times |\mathcal{A}_{\widehat{\mathbb{E}}}|}}$ such that each row $G_{\ast}^{(t)}$ is a multi-hot vector representing a distinct timestep $t$, with $1$'s in the columns that correspond to the gold actions. This conversion is done such that the $\shift$ and $\eoa$ actions always occupy different time steps, but $\transition{}$ and $\reduce{}$ actions of different entity types can coexist at the same time step. Then, $\mathbf{G}$ is changed during the action generation cycle by incorporating the probabilities of the model's decisions on $\transition{}$ and $\reduce{}$ actions from other datasets. More concretely, at time step $t$ of the cycle, for $a_i\in \mathcal{A}_{\widehat{\mathbb{E}}}\setminus \mathcal{A}_{\mathbb{E}_i}$, $G_{a_i}^{(t)}$ is set to be equal to $\sigma\left(u_{a_i}^{(t)}\right)$, where $\sigma$ is the sigmoid function. In addition, when $G_{\shift}^{(t)}=1$ and $u_{a_i}^{(t)} > u_{\shift}^{(t)}$ for some $a_i\in \mathcal{A}_{\widehat{\mathbb{E}}}\setminus \mathcal{A}_{\mathbb{E}_i}$, that is, when the model is trying to open/close a new span of an entity type of other dataset $D_j$ ($j\neq i$), the value $G_{\shift}^{(t)}$ is changed to $\sigma\left(u_{\shift}^{(t)}\right)$. In this case, a one-hot vector is inserted in $\mathbf{G}$ after $G_{*}^{(t)}$, so that, on the next time step $t+1$, $G_{\shift}^{(t+1)}=1$ and $G_{a_i}^{(t+1)}=0$ for all $a_i\in \widehat{\mathbb{E}}\setminus\{\shift\}$. This procedure ensures that $\mathbf{G}$ still reflects the original gold annotations in the columns corresponding to $\transition{}$ and $\reduce{}$ actions of entity types in the source dataset, but incorporates the model's probabilities for other actions. Then, by setting, for each $1\leq t \leq T_{\text{final}}$,

\begin{align*}
\begin{split}
L^{(t)} =& -\frac{1}{|\mathcal{A}_{\widehat{\mathbb{E}}}|}\, \sum_{a_i\in \mathcal{A}_{\widehat{\mathbb{E}}}}
\Bigg(
G_{a_i}^{(t)} \ \log\left(\sigma\left(u_{a_i}^{(t)}\right)\right) \\ 
& + \left(1-G_{a_i}^{(t)}\right)\ \log\left(1-\sigma\left(u_{a_i}^{(t)}\right)\right)\Bigg)
\end{split}
\end{align*}
 
\noindent the total loss of the sample is computed as

\begin{equation*}
    L = \frac{1}{T_{\text{final}}}\sum_{t=1}^{T_{\text{final}}}L^{(t)}\ .
\end{equation*}

\noindent Given how $\mathbf{G}$ is constructed, this ensures the aforementioned constraints i) and ii) on the loss function are satisfied.

\section{Experiments and Results}
To evaluate the performance of the proposed architecture for the NER task, single-task experiments were conducted on benchmarks datasets, specifically the English subset of CoNLL-2003 \cite{conll_tjong-kim-sang-de-meulder-2003-introduction} and GENIA \cite{genia_10.1093/bioinformatics/btg1023}. The model's multi-task performance is also assessed by training it with an ensemble of six biomedical datasets that have been extensively used in previous research.
In order to demonstrate the viability of SRU-NER for downstream applications, a model is evaluated in a cross-corpus setting by replicating the experimental setup of \citealp{HunFlair2}. Finally, two further experiments are conducted to quantify the reliability of the multi-task models' predictions for entity types that are not explicitly annotated in the test corpora, providing a more comprehensive assessment of their generalization capabilities.
\par The datasets used across the following sections and respective experimental setup are described in Appendix~\ref{appendix:datasets}. Training details can be found in Appendix~\ref{appendix:training}. For evaluation purposes, a predicted mention is considered a true positive if and only if both its span boundaries and entity type exactly match the gold annotation. Results are reported for each dataset using mention-level micro F1 scores.

\subsection{Single-task performance}
\label{sec:ST_results}
\par The results of the two single-task models are presented in Table \ref{tab:results_benchmarks}. The proposed model achieves micro F1 scores of 94.48\% on the CoNLL-2003 dataset, and 80.10\% on the GENIA dataset. These results are very close to state-of-the-art (SOTA), demonstrating the competitiveness of SRU-NER in both flat and nested NER scenarios. Nonetheless, in contrast to our approach, the models presented as SOTA were trained using both the training and development splits of their respective datasets. This difference in training data availability may contribute to the observed performance gap, particularly on GENIA, where additional annotated data could provide further benefits in capturing complex biomedical terminology.

\begin{table}[h]
\small
    \centering
    \begin{tabular}{lrl}
    \toprule
Dataset  & \multicolumn{1}{c}{SRU-NER} & \multicolumn{1}{c}{SOTA} \\ \midrule
CoNLL    & 94.48                         & 94.6*, \cite{ACE_wang-etal-2021-automated}                    \\
GENIA & 80.10                         & 81.53*, \cite{shen-etal-2023-diffusionner}                    \\ \bottomrule                                         
\end{tabular}
\caption{Micro-F1 scores of single-task models on benchmark datasets. The entity counts of the datasets can be found in Table \ref{tab:ST_entity_counts}. The * symbol indicates that a model was trained on both the training and development splits of the corpus.
}
\label{tab:results_benchmarks}
\end{table}

\begin{table}[h]
    \centering
    \small
    \begin{tabular}{lrl}
    \toprule
Dataset  & \multicolumn{1}{c}{SRU-NER} & \multicolumn{1}{c}{SOTA} \\ \midrule
BC2GM    & 85.43                         & 85.48*~\cite{sun_MRC}                    \\
BC4CHEMD & 92.64                         & 92.92*~\cite{sun_MRC}                    \\
BC5CDR   & 90.61                         & 91.90~\citep{zhang2023optimizing}                     \\
JNLPBA   & 77.12                         & 78.93*~\cite{sun_MRC}                    \\
Linnaeus & 89.62                         & 94.13~\citep{Habibi_Linnaeus}                     \\
NCBI Disease    & 89.25                         & 90.04*~\cite{sun_MRC}              \\ \midrule
\textit{Average} & 87.45 & \\
\bottomrule                                            
\end{tabular}
\caption{Micro-F1 scores of single-task models trained on the datasets used for the multi-task model described in section \ref{sec:results_MT}. SOTA results are for single-task models. The * symbol indicates the model was trained on a larger training split.}
\label{tab:ST_results_for_MT}
\end{table}

\subsection{Multi-task performance}
\label{sec:results_MT}

In Table~\ref{tab:MT_results}, we show the results of SRU-NER trained on an ensemble $\{D_i\}_{i=1}^6$ of six biomedical datasets, annotated for $\left|\cup_i\, \mathbb{E}_i\right|=8$ entity types. Since there are entity types which are annotated on more than one dataset (\textit{e.g.} BC4CHEMD and BC5CDR are both annotated with mentions of the Chemical type), two evaluation scenarios are considered, that differ in how these type labels are interpreted. Recalling that the model infers mentions with entity types in the disjoint union set $\widehat{\mathbb{E}}=\sqcup_i\, \mathbb{E}_i$, given a sentence coming from the test split of dataset $D_i$ of the ensemble, in the case of:
\begin{enumerate}[i)]
    \item \textit{disjoint evaluation}, the predicted spans of types $\mathbb{E}_i \subset \widehat{\mathbb{E}}$ are compared against the gold ones, and any predicted span of type in $\widehat{\mathbb{E}}\setminus\mathbb{E}_i$ is discarded;
    \item \textit{merged evaluation}, the entity types of predicted spans are mapped to $\cup_i\, \mathbb{E}_i$, and the spans whose mapped types do not also belong in $\mathbb{E}_i$ are discarded; the remaining spans are compared against the gold ones.
\end{enumerate}

\begin{figure}[h]
    \centering
    \includegraphics[width=\linewidth]{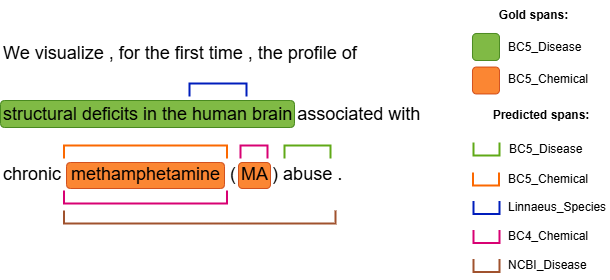}
    \caption{Example of a sentence from the test split of the BC5CDR corpus \cite{dataset:BC5CDR}, together with gold spans and predicted spans as annotated by the MTL model described in section \ref{sec:results_MT}. The model is trained on six datasets, covering eight entity types $\cup_i\,\mathbb{E}_i =\{\text{Chemical}, \text{Disease}, \ldots\}$. Notice that some of these types are common to multiple datasets (namely, 'Chemical', annotated on both the BC4CHEMD and BC5CDR datasets; and 'Disease', annotated on both the BC5CDR and NCBI datasets). SRU-NER tags spans with one of 11 possible types, built by adjoining the dataset name to the original type name, such that ${\widehat{\mathbb{E}}=\{\text{BC4\_Chemical}, \text{BC5\_Chemical},\ldots\}}$. In the \textit{disjoint} evaluation case, and since this sentence comes from the BC5CDR corpus, metrics are computed by considering only the spans whose types in $\widehat{\mathbb{E}}$ start with the BC5 shorthand, resulting in one true positive, one false positive and two false negatives. In the \textit{merged} evaluation case, spans whose types in $\widehat{\mathbb{E}}$ do not end with 'Chemical' or 'Disease' are discarded, and the remaining spans have their types mapped to $\cup_i\,\mathbb{E}_i$ by removing the dataset identifier. With these spans, there are two true positives, two false positives and one false negative in the sentence.}
    \label{fig:spans}
\end{figure}

\noindent An example of the predictions of the model on a test sentence, together with which spans are used to compute the metrics on the two evaluation scenarios is present in Figure \ref{fig:spans}.

Compared to previous MTL models, the proposed model achieves the best or second-best F1 scores in the disjoint evaluation setting. These results are obtained without relying on task-specific classification layers~\citep{cross_type_wang, khan2020mtbionermultitasklearningbiomedical} or training multiple single-task teacher models followed by knowledge distillation into a student model~\citep{taughtnet}. Instead, a single unified model learns each task directly from its respective annotated dataset while preserving the performance of other tasks. This approach enables joint decoding, thereby eliminating the need for post-processing steps to resolve span conflicts.


Table~\ref{tab:ST_results_for_MT} presents F1 scores for single-task models trained on each dataset used in the multi-task setting, alongside SOTA references. The results demonstrate that the proposed model remains competitive in the single-task setting.
The average F1 score of the six single-task SRU-NER models is 0.82 percentage points higher than the dataset-average F1 of the multi-task SRU-NER model under the disjoint evaluation setting. This aligns with previous findings, which suggest that while multi-task training improves model robustness across datasets, it may lead to lower in-corpus performance compared to single-task models \cite{YIN2024104731}. To further investigate the generalization capabilities of the model, the next section presents an evaluation in a cross-corpus setting.

\begin{table}[h]
    \centering
    \small
\begin{tabular}{@{}llrr@{}}

\toprule
Dataset                      & Entity type & SRU-NER & Baseline \\ \midrule
BioID                        & Species     & \textbf{62.41}     & 58.21     \\
\multirow[t]{2}{*}{MedMentions} & Chemical    & \textbf{59.53}     & 58.40     \\
                             & Disease     & \textbf{62.48}     & 62.18     \\
tmVar3                       & Gene        & \textbf{90.38}     & 87.87     \\ \midrule
\multicolumn{2}{c}{\textit{Average}}       & \textbf{68.70}     & 66.67     \\ \bottomrule
\end{tabular}
\caption{Mention-level F1 scores for the cross-corpus experiment. SRU-NER was trained on an emsemble of 8 biomedical datasets, and evaluated on 3 independent corpora. Baseline refers to the scores obtained by \cite{HunFlair2}. Best scores are in \textbf{bold}.}
\label{tab:cross_eval_results}
\end{table}

\begin{table}[ht]
\centering
\small
\begin{tabular}{lcc}
\toprule
Training datasets & Chemical & Disease \\ \midrule
Only BC5-Chemical & 91.27    & ---     \\
Only BC5-Disease  & ---      & 85.41   \\
Both              & \textbf{91.81}    & \textbf{86.10}   \\ \bottomrule
\end{tabular}
\caption{Global prediction F1 scores on the test split of BC5CDR of models trained on synthetic datasets. Best scores are \textbf{bold}.}
\label{tab:simulated}
\end{table}

\begin{table*}[t!]
\small
\centering
\begin{tabular}{@{}lccccccccc@{}}
\toprule
\multirow{2}{*}{Entity} & \multicolumn{3}{c}{\textbf{SRU-NER-CoNLL}} & \multicolumn{3}{c}{\textbf{SRU-NER-BC5}} & \multicolumn{3}{c}{\textbf{SRU-NER-MTL}} \\
                        & \textit{P}               & \textit{R}              & \textit{F1}             & \textit{P}               & \textit{R}              & \textit{F1}             & \textit{P}           & \textit{R}         & \textit{F1}        \\ \midrule
Chemical                & 24.71           & 87.76          & 38.57          & ---             & ---            & ---            & 75.00       & 9.18      & 16.36     \\
Disease                 & 25.25           & 83.33          & 38.76          & ---             & ---            & ---            & 88.46       & 38.33     & 53.49     \\
LOC                     & ---             & ---            & ---            & 98.25           & 88.89          & 93.33          & 100.00      & 96.83     & 98.39     \\
ORG                     & ---             & ---            & ---            & 80.00           & 80.00          & 80.00          & 86.36       & 71.25     & 78.08     \\
PER                     & ---             & ---            & ---            & 94.44           & 94.44          & 94.44          & 100.00      & 22.22     & 36.36    \\ \bottomrule
\end{tabular}
\caption{Human evaluation of the out-of-domain predictions made by three models. \textit{P} stands for precision, \textit{R} for the simulated recall, and \textit{F1} for the F1 computed with the former two metrics. Details on how these metrics were computed can be found in Appendix \ref{appendix:human_eval}.}
\label{tab:human_eval}
\end{table*}

\vspace{-2mm}

\subsection{Cross-corpus evaluation}
\label{sec:results_CC}

Table \ref{tab:cross_eval_results} presents the results of the proposed model in a cross-corpus evaluation, replicating the experimental setup of \citealp{HunFlair2}. The model was trained on an ensemble of nine datasets covering five entity types and evaluated on three independent corpora annotated for four of these types. The results indicate that SRU-NER outperforms the baseline by an average of 2.03\%, with notable improvements for the Species (4.2\%) and Gene (2.51\%) entity types. These findings underscore the robustness of the model and demonstrate its potential for downstream applications. For reference, in-corpus F1 scores are provided in Appendix~\ref{appendix:CC}.

\subsection{Evaluation of global predictions}
\label{sec:global_predictions}
The previous experiments evaluated the model's \textit{local} prediction ability. Specifically, when the model is trained on a collection $\{D_i\}_{i=1}^K$, where each dataset $D_i$ was annotated for entity types $\mathbb{E}_i$, its performance was assessed on a test dataset $D_{\text{test}}$ annotated with entity types $\mathbb{E}_{\text{test}} \subseteq \mathbb{E}_j$ for some $j\in \{1,\ldots,K\}$. However, the model generates predictions for spans of all entity types in $\cup_i\, \mathbb{E}_i$ within $D_{\text{test}}$. To evaluate its global prediction ability, it is necessary to test the model on a dataset annotated with a superset of entity types spanning multiple training datasets.

First, following the approach of \citealp{huang-etal-2019-learning}, a synthetic dataset is constructed from the BC5CDR corpus. The original training set is randomly partitioned into two disjoint subsets: one containing only Chemical annotations (\textit{BC5-Chemical}) and another containing only Disease annotations (\textit{BC5-Disease}). Additional details on these synthetic datasets are provided in Appendix \ref{appendix:datasets}. Two single-task models are trained separately on each subset, while a multi-task model is trained on both. All models are evaluated on the original test split of the BC5CDR corpus. The results, presented in Table \ref{tab:simulated}, demonstrate that the training strategy outlined in section \ref{sec:training_strategy} effectively enables the model to make accurate global predictions across entity types from different training datasets.

Secondly, a multi-task model is trained on both the CoNLL-2003 dataset and the BC5CDR dataset. This approach results in a model capable of recognizing six entity types: four from the general domain (\textit{LOC}, \textit{MISC}, \textit{ORG}, \textit{PER}) and two from the biomedical domain (\textit{Chemical}, \textit{Disease}). To assess the model’s ability to generalize across domains, its predictions of general-domain entity types in the test split of the BC5CDR dataset and, conversely, its predictions of biomedical entity types in the test split of the CoNLL dataset are evaluated. The results of the multi-task model can be found in Table~\ref{tab:human_eval} under the column SRU-NER-MTL. Since gold annotations for these cross-domain predictions are not available, the evaluation was conducted manually by two human annotators. Provided with definitions of the entity types, they independently assessed whether the model’s predictions were correct. This human evaluation was also conducted for the predictions of two single-task models: one trained on CoNLL-2003 and evaluated on the BC5CDR test set (SRU-NER-CoNLL), and another trained on BC5CDR and evaluated on the CoNLL-2003 test set (SRU-NER-BC5). A comparison between the single-task and multi-task models reveals that multi-task SRU-NER is, on average, 25.4\% more precise in identifying out-of-domain spans. For instance, the single-task model trained on biomedical entity types incorrectly classified \textit{lead} as a chemical in the CoNLL-2003 sentence: \textit{"Indonesian keeper Hendro Kartiko produced a string of fine saves to prevent the Koreans increasing their lead."} In contrast, the multi-task model did not make this error. Further details on this experiment are provided in Appendix \ref{appendix:human_eval}.

\section{Conclusion}

This work presents SRU-NER, a novel architecture for Named Entity Recognition capable of handling nested entities through a transition-based parsing approach. The model integrates a Slot-based Recurrent Unit (SRU) to maintain an evolving representation of past actions, enabling effective entity extraction. Unlike traditional multi-task learning approaches that rely on separate models for different entity types, SRU-NER employs a unified learning strategy, allowing a single model to learn from multiple datasets. This approach improves adaptability to annotation inconsistencies and enhances generalization across domains.

Experimental results demonstrate that SRU-NER achieves strong performance in both single- and multi-task settings, with cross-corpus evaluations and human assessments confirming the robustness of its predictions. These findings highlight the advantages of training a single multi-task model for BioNER and suggest promising directions for future research, including advancements in nested entity recognition and domain adaptability.

\section*{Limitations}

While the proposed SRU-NER architecture has demonstrated effectiveness for named entity recognition in general and biomedical domains, its performance in other domains, such as legal or financial, was not evaluated. Furthermore, the generalizability of the findings may be limited, as evaluations on community-available biomedical datasets may not fully capture the diversity of real-world biomedical text. Finally, the assessment of global prediction ability in a cross-domain scenario relied on human annotators, introducing a degree of subjectivity to the evaluation.
\par While the model achieves competitive results, we note that no extensive hyperparameter search was conducted. A more systematic tuning of hyperparameters could potentially yield further improvements. Additionally, the training strategy presents opportunities for refinement, notably in the sampling strategy utilized within the multi-task learning framework.

\section*{Acknowledgments}
This research was supported by the Portuguese
Recovery and Resilience Plan through project
C645008882-00000055 (\textit{i.e.}, the Center For Responsible AI).

\bibliography{acl_latex}

\begin{thebibliography}{49}
\providecommand{\natexlab}[1]{#1}

\bibitem[{Arighi et~al.(2017)Arighi, Hirschman, Lemberger, Bayer, Liechti, Comeau, and Wu}]{dataset:BioID}
Cecilia Arighi, Lynette Hirschman, Thomas Lemberger, Samuel Bayer, Robin Liechti, Donald Comeau, and Cathy Wu. 2017.
\newblock Bio-id track overview.
\newblock In \emph{BioCreative VI Challenge Evaluation Workshop}, volume 482, page 376.

\bibitem[{Banerjee et~al.(2021)Banerjee, Pal, Devarakonda, and Baral}]{banerjee_biomedical_2021}
Pratyay Banerjee, Kuntal~Kumar Pal, Murthy Devarakonda, and Chitta Baral. 2021.
\newblock \href {https://doi.org/10.1145/3465221} {Biomedical {Named} {Entity} {Recognition} via {Knowledge} {Guidance} and {Question} {Answering}}.
\newblock \emph{ACM Trans. Comput. Healthcare}, 2(4):33:1--33:24.

\bibitem[{Collier et~al.(2004)Collier, Ohta, Tsuruoka, Tateisi, and Kim}]{dataset:JNLPBA}
Nigel Collier, Tomoko Ohta, Yoshimasa Tsuruoka, Yuka Tateisi, and Jin-Dong Kim. 2004.
\newblock \href {https://aclanthology.org/W04-1213/} {Introduction to the bio-entity recognition task at {JNLPBA}}.
\newblock In \emph{Proceedings of the International Joint Workshop on Natural Language Processing in Biomedicine and its Applications ({NLPBA}/{B}io{NLP})}, pages 73--78, Geneva, Switzerland. COLING.

\bibitem[{Conneau et~al.(2020)Conneau, Khandelwal, Goyal, Chaudhary, Wenzek, Guzm{\'a}n, Grave, Ott, Zettlemoyer, and Stoyanov}]{conneau-etal-2020-unsupervised}
Alexis Conneau, Kartikay Khandelwal, Naman Goyal, Vishrav Chaudhary, Guillaume Wenzek, Francisco Guzm{\'a}n, Edouard Grave, Myle Ott, Luke Zettlemoyer, and Veselin Stoyanov. 2020.
\newblock \href {https://doi.org/10.18653/v1/2020.acl-main.747} {Unsupervised cross-lingual representation learning at scale}.
\newblock In \emph{Proceedings of the 58th Annual Meeting of the Association for Computational Linguistics}, pages 8440--8451, Online. Association for Computational Linguistics.

\bibitem[{Crichton et~al.(2017)Crichton, Pyysalo, Chiu, and Korhonen}]{crichton_neural_2017}
Gamal Crichton, Sampo Pyysalo, Billy Chiu, and Anna Korhonen. 2017.
\newblock \href {https://doi.org/10.1186/s12859-017-1776-8} {A neural network multi-task learning approach to biomedical named entity recognition}.
\newblock \emph{BMC Bioinformatics}, 18(1):368.

\bibitem[{Do{\u g}an et~al.(2014)Do{\u g}an, Leaman, and Lu}]{dataset:NCBIDisease}
Rezarta~Islamaj Do{\u g}an, Robert Leaman, and Zhiyong Lu. 2014.
\newblock {NCBI} disease corpus: a resource for disease name recognition and concept normalization.
\newblock \emph{J Biomed Inform}, 47:1--10.

\bibitem[{Dyer et~al.(2015)Dyer, Ballesteros, Ling, Matthews, and Smith}]{dyer-etal-2015-transition}
Chris Dyer, Miguel Ballesteros, Wang Ling, Austin Matthews, and Noah~A. Smith. 2015.
\newblock \href {https://doi.org/10.3115/v1/P15-1033} {Transition-based dependency parsing with stack long short-term memory}.
\newblock In \emph{Proceedings of the 53rd Annual Meeting of the Association for Computational Linguistics and the 7th International Joint Conference on Natural Language Processing (Volume 1: Long Papers)}, pages 334--343, Beijing, China. Association for Computational Linguistics.

\bibitem[{Ganea and Hofmann(2017)}]{ganea-hofmann-2017-deep}
Octavian-Eugen Ganea and Thomas Hofmann. 2017.
\newblock \href {https://doi.org/10.18653/v1/D17-1277} {Deep joint entity disambiguation with local neural attention}.
\newblock In \emph{Proceedings of the 2017 Conference on Empirical Methods in Natural Language Processing}, pages 2619--2629, Copenhagen, Denmark. Association for Computational Linguistics.

\bibitem[{Gerner et~al.(2010)Gerner, Nenadic, and Bergman}]{dataset:Linnaeus}
Martin Gerner, Goran Nenadic, and Casey~M. Bergman. 2010.
\newblock \href {https://doi.org/10.1186/1471-2105-11-85} {Linnaeus: A species name identification system for biomedical literature}.
\newblock \emph{BMC Bioinformatics}, 11(1):85.

\bibitem[{Greenberg et~al.(2018)Greenberg, Bansal, Verga, and McCallum}]{greenberg-etal-2018-marginal}
Nathan Greenberg, Trapit Bansal, Patrick Verga, and Andrew McCallum. 2018.
\newblock \href {https://doi.org/10.18653/v1/D18-1306} {Marginal likelihood training of {B}i{LSTM}-{CRF} for biomedical named entity recognition from disjoint label sets}.
\newblock In \emph{Proceedings of the 2018 Conference on Empirical Methods in Natural Language Processing}, pages 2824--2829, Brussels, Belgium. Association for Computational Linguistics.

\bibitem[{Gurulingappa et~al.(2010)Gurulingappa, Klinger, Hofmann-Apitius, and Fluck}]{dataset:SCAIdisease}
Harsha Gurulingappa, Roman Klinger, Martin Hofmann-Apitius, and Juliane Fluck. 2010.
\newblock \href {http://www.nactem.ac.uk/biotxtm/papers/Gurulingappa.pdf} {An empirical evaluation of resources for the identification of diseases and adverse effects in biomedical literature}.
\newblock In \emph{{2nd Workshop on Building and evaluating resources for biomedical text mining (7th edition of the Language Resources and Evaluation Conference)}}, Valetta, Malta.

\bibitem[{Habibi et~al.(2017)Habibi, Weber, Neves, Wiegandt, and Leser}]{Habibi_Linnaeus}
Maryam Habibi, Leon Weber, Mariana Neves, David~Luis Wiegandt, and Ulf Leser. 2017.
\newblock \href {https://doi.org/10.1093/bioinformatics/btx228} {Deep learning with word embeddings improves biomedical named entity recognition}.
\newblock \emph{Bioinformatics}, 33(14):i37--i48.

\bibitem[{Huang et~al.(2019)Huang, Dong, Boschee, and Peng}]{huang-etal-2019-learning}
Xiao Huang, Li~Dong, Elizabeth Boschee, and Nanyun Peng. 2019.
\newblock \href {https://doi.org/10.18653/v1/K19-1048} {Learning a unified named entity tagger from multiple partially annotated corpora for efficient adaptation}.
\newblock In \emph{Proceedings of the 23rd Conference on Computational Natural Language Learning (CoNLL)}, pages 515--527, Hong Kong, China. Association for Computational Linguistics.

\bibitem[{Islamaj et~al.(2021{\natexlab{a}})Islamaj, Leaman, Kim, Kwon, Wei, Comeau, Peng, Cissel, Coss, Fisher, Guzman, Kochar, Koppel, Trinh, Sekiya, Ward, Whitman, Schmidt, and Lu}]{dataset:NLMchem}
Rezarta Islamaj, Robert Leaman, Sun Kim, Dongseop Kwon, Chih-Hsuan Wei, Donald~C Comeau, Yifan Peng, David Cissel, Cathleen Coss, Carol Fisher, Rob Guzman, Preeti~Gokal Kochar, Stella Koppel, Dorothy Trinh, Keiko Sekiya, Janice Ward, Deborah Whitman, Susan Schmidt, and Zhiyong Lu. 2021{\natexlab{a}}.
\newblock {NLM-Chem}, a new resource for chemical entity recognition in {PubMed} full text literature.
\newblock \emph{Sci Data}, 8(1):91.

\bibitem[{Islamaj et~al.(2021{\natexlab{b}})Islamaj, Wei, Cissel, Miliaras, Printseva, Rodionov, Sekiya, Ward, and Lu}]{dataset:NLMGene}
Rezarta Islamaj, Chih-Hsuan Wei, David Cissel, Nicholas Miliaras, Olga Printseva, Oleg Rodionov, Keiko Sekiya, Janice Ward, and Zhiyong Lu. 2021{\natexlab{b}}.
\newblock {NLM-Gene}, a richly annotated gold standard dataset for gene entities that addresses ambiguity and multi-species gene recognition.
\newblock \emph{J Biomed Inform}, 118:103779.

\bibitem[{Keraghel et~al.(2024)Keraghel, Morbieu, and Nadif}]{keraghel2024recentadvancesnamedentity}
Imed Keraghel, Stanislas Morbieu, and Mohamed Nadif. 2024.
\newblock \href {https://arxiv.org/abs/2401.10825} {Recent advances in named entity recognition: A comprehensive survey and comparative study}.
\newblock \emph{Preprint}, arXiv:2401.10825.

\bibitem[{Khan et~al.(2020)Khan, Ziyadi, and AbdelHady}]{khan2020mtbionermultitasklearningbiomedical}
Muhammad~Raza Khan, Morteza Ziyadi, and Mohamed AbdelHady. 2020.
\newblock \href {https://arxiv.org/abs/2001.08904} {Mt-bioner: Multi-task learning for biomedical named entity recognition using deep bidirectional transformers}.
\newblock \emph{Preprint}, arXiv:2001.08904.

\bibitem[{Kim et~al.(2003)Kim, Ohta, Tateisi, and Tsujii}]{genia_10.1093/bioinformatics/btg1023}
J.-D. Kim, T.~Ohta, Y.~Tateisi, and J.~Tsujii. 2003.
\newblock \href {https://doi.org/10.1093/bioinformatics/btg1023} {Genia corpus—a semantically annotated corpus for bio-textmining}.
\newblock \emph{Bioinformatics}, 19:i180--i182.

\bibitem[{Kolarik et~al.(2008)Kolarik, Klinger, Friedrich, Hofmann-Apitius, and Fluck}]{dataset:SCAIchemical}
Corinna Kolarik, Roman Klinger, Christoph~M. Friedrich, Martin Hofmann-Apitius, and Juliane Fluck. 2008.
\newblock \href {http://www.romanklinger.de/publications/kolarik2008.pdf} {{Chemical Names: Terminological Resources and Corpora Annotation}}.
\newblock In \emph{{Workshop on Building and evaluating resources for biomedical text mining (6th edition of the Language Resources and Evaluation Conference)}}, pages 51--58, Marrakech, Morocco.

\bibitem[{Krallinger et~al.(2015)Krallinger, Rabal, Leitner, Vazquez, Salgado, Lu, Leaman, Lu, Ji, Lowe, Sayle, Batista-Navarro, Rak, Huber, Rockt{\"a}schel, Matos, Campos, Tang, Xu, Munkhdalai, Ryu, Ramanan, Nathan, {\v{Z}}itnik, Bajec, Weber, Irmer, Akhondi, Kors, Xu, An, Sikdar, Ekbal, Yoshioka, Dieb, Choi, Verspoor, Khabsa, Giles, Liu, Ravikumar, Lamurias, Couto, Dai, Tsai, Ata, Can, Usi{\'e}, Alves, Segura-Bedmar, Mart{\'i}nez, Oyarzabal, and Valencia}]{dataset:BC4CHEMD}
Martin Krallinger, Obdulia Rabal, Florian Leitner, Miguel Vazquez, David Salgado, Zhiyong Lu, Robert Leaman, Yanan Lu, Donghong Ji, Daniel~M. Lowe, Roger~A. Sayle, Riza~Theresa Batista-Navarro, Rafal Rak, Torsten Huber, Tim Rockt{\"a}schel, S{\'e}rgio Matos, David Campos, Buzhou Tang, Hua Xu, Tsendsuren Munkhdalai, Keun~Ho Ryu, S.~V. Ramanan, Senthil Nathan, Slavko {\v{Z}}itnik, Marko Bajec, Lutz Weber, Matthias Irmer, Saber~A. Akhondi, Jan~A. Kors, Shuo Xu, Xin An, Utpal~Kumar Sikdar, Asif Ekbal, Masaharu Yoshioka, Thaer~M. Dieb, Miji Choi, Karin Verspoor, Madian Khabsa, C.~Lee Giles, Hongfang Liu, Komandur~Elayavilli Ravikumar, Andre Lamurias, Francisco~M. Couto, Hong-Jie Dai, Richard Tzong-Han Tsai, Caglar Ata, Tolga Can, Anabel Usi{\'e}, Rui Alves, Isabel Segura-Bedmar, Paloma Mart{\'i}nez, Julen Oyarzabal, and Alfonso Valencia. 2015.
\newblock \href {https://doi.org/10.1186/1758-2946-7-S1-S2} {The chemdner corpus of chemicals and drugs and its annotation principles}.
\newblock \emph{Journal of Cheminformatics}, 7(1):S2.

\bibitem[{Li et~al.(2016)Li, Sun, Johnson, Sciaky, Wei, Leaman, Davis, Mattingly, Wiegers, and Lu}]{dataset:BC5CDR}
Jiao Li, Yueping Sun, Robin~J Johnson, Daniela Sciaky, Chih-Hsuan Wei, Robert Leaman, Allan~Peter Davis, Carolyn~J Mattingly, Thomas~C Wiegers, and Zhiyong Lu. 2016.
\newblock {BioCreative} {V} {CDR} task corpus: a resource for chemical disease relation extraction.
\newblock \emph{Database (Oxford)}, 2016.

\bibitem[{Li et~al.(2022)Li, Sun, Han, and Li}]{survey_li}
Jing Li, Aixin Sun, Jianglei Han, and Chenliang Li. 2022.
\newblock \href {https://doi.org/10.1109/TKDE.2020.2981314} {A survey on deep learning for named entity recognition}.
\newblock \emph{IEEE Transactions on Knowledge and Data Engineering}, 34(1):50--70.

\bibitem[{Luo et~al.(2022)Luo, Lai, Wei, Arighi, and Lu}]{dataset:BioRED}
Ling Luo, Po-Ting Lai, Chih-Hsuan Wei, Cecilia~N Arighi, and Zhiyong Lu. 2022.
\newblock \href {https://doi.org/10.1093/bib/bbac282} {Biored: a rich biomedical relation extraction dataset}.
\newblock \emph{Briefings in Bioinformatics}, 23(5):bbac282.

\bibitem[{Luo et~al.(2023)Luo, Wei, Lai, Leaman, Chen, and Lu}]{aioner}
Ling Luo, Chih-Hsuan Wei, Po-Ting Lai, Robert Leaman, Qingyu Chen, and Zhiyong Lu. 2023.
\newblock \href {https://doi.org/10.1093/bioinformatics/btad310} {{AIONER: all-in-one scheme-based biomedical named entity recognition using deep learning}}.
\newblock \emph{Bioinformatics}, 39(5):btad310.

\bibitem[{Marinho et~al.(2019)Marinho, Mendes, Miranda, and Nogueira}]{marinho-etal-2019-hierarchical}
Zita Marinho, Afonso Mendes, Sebasti{\~a}o Miranda, and David Nogueira. 2019.
\newblock \href {https://doi.org/10.18653/v1/W19-1904} {Hierarchical nested named entity recognition}.
\newblock In \emph{Proceedings of the 2nd Clinical Natural Language Processing Workshop}, pages 28--34, Minneapolis, Minnesota, USA. Association for Computational Linguistics.

\bibitem[{Mehmood et~al.(2019)Mehmood, Gerevini, Lavelli, and Serina}]{10.1007/978-3-030-35166-3_31}
Tahir Mehmood, Alfonso Gerevini, Alberto Lavelli, and Ivan Serina. 2019.
\newblock Leveraging multi-task learning for biomedical named entity recognition.
\newblock In \emph{AI*IA 2019 -- Advances in Artificial Intelligence}, pages 431--444, Cham. Springer International Publishing.

\bibitem[{Mohan and Li(2019)}]{dataset:MedMentions}
Sunil Mohan and Donghui Li. 2019.
\newblock \href {https://arxiv.org/abs/1902.09476} {Medmentions: A large biomedical corpus annotated with umls concepts}.
\newblock \emph{Preprint}, arXiv:1902.09476.

\bibitem[{Moscato et~al.(2023)Moscato, Postiglione, Sansone, and Sperlí}]{taughtnet}
Vincenzo Moscato, Marco Postiglione, Carlo Sansone, and Giancarlo Sperlí. 2023.
\newblock \href {https://doi.org/10.1109/JBHI.2023.3244044} {Taughtnet: Learning multi-task biomedical named entity recognition from single-task teachers}.
\newblock \emph{IEEE Journal of Biomedical and Health Informatics}, 27(5):2512--2523.

\bibitem[{Pafilis et~al.(2013)Pafilis, Frankild, Fanini, Faulwetter, Pavloudi, Vasileiadou, Arvanitidis, and Jensen}]{dataset:S800}
Evangelos Pafilis, Sune~P Frankild, Lucia Fanini, Sarah Faulwetter, Christina Pavloudi, Aikaterini Vasileiadou, Christos Arvanitidis, and Lars~Juhl Jensen. 2013.
\newblock The {SPECIES} and {ORGANISMS} resources for fast and accurate identification of taxonomic names in text.
\newblock \emph{PLoS One}, 8(6):e65390.

\bibitem[{Park et~al.(2024)Park, Son, and Rho}]{review_flat_nested}
Yesol Park, Gyujin Son, and Mina Rho. 2024.
\newblock \href {https://doi.org/10.3390/app14209302} {Biomedical flat and nested named entity recognition: Methods, challenges, and advances}.
\newblock \emph{Applied Sciences}, 14(20).

\bibitem[{Sharma et~al.(2022)Sharma, Amrita, Chakraborty, and Kumar}]{10.1007}
Abhishek Sharma, Amrita, Sudeshna Chakraborty, and Shivam Kumar. 2022.
\newblock Named entity recognition in natural language processing: A systematic review.
\newblock In \emph{Proceedings of Second Doctoral Symposium on Computational Intelligence}, pages 817--828, Singapore. Springer Singapore.

\bibitem[{Shen et~al.(2021)Shen, Ma, Tan, Zhang, Wang, and Lu}]{shen-etal-2021-locate}
Yongliang Shen, Xinyin Ma, Zeqi Tan, Shuai Zhang, Wen Wang, and Weiming Lu. 2021.
\newblock \href {https://doi.org/10.18653/v1/2021.acl-long.216} {Locate and label: A two-stage identifier for nested named entity recognition}.
\newblock In \emph{Proceedings of the 59th Annual Meeting of the Association for Computational Linguistics and the 11th International Joint Conference on Natural Language Processing (Volume 1: Long Papers)}, pages 2782--2794, Online. Association for Computational Linguistics.

\bibitem[{Shen et~al.(2023)Shen, Song, Tan, Li, Lu, and Zhuang}]{shen-etal-2023-diffusionner}
Yongliang Shen, Kaitao Song, Xu~Tan, Dongsheng Li, Weiming Lu, and Yueting Zhuang. 2023.
\newblock \href {https://doi.org/10.18653/v1/2023.acl-long.215} {{D}iffusion{NER}: Boundary diffusion for named entity recognition}.
\newblock In \emph{Proceedings of the 61st Annual Meeting of the Association for Computational Linguistics (Volume 1: Long Papers)}, pages 3875--3890, Toronto, Canada. Association for Computational Linguistics.

\bibitem[{Shen et~al.(2022)Shen, Wang, Tan, Xu, Xie, Huang, Lu, and Zhuang}]{PIQN_shen-etal-2022-parallel}
Yongliang Shen, Xiaobin Wang, Zeqi Tan, Guangwei Xu, Pengjun Xie, Fei Huang, Weiming Lu, and Yueting Zhuang. 2022.
\newblock \href {https://doi.org/10.18653/v1/2022.acl-long.67} {Parallel instance query network for named entity recognition}.
\newblock In \emph{Proceedings of the 60th Annual Meeting of the Association for Computational Linguistics (Volume 1: Long Papers)}, pages 947--961, Dublin, Ireland. Association for Computational Linguistics.

\bibitem[{Smith et~al.(2008)Smith, Tanabe, Ando, Kuo, Chung, Hsu, Lin, Klinger, Friedrich, Ganchev, Torii, Liu, Haddow, Struble, Povinelli, Vlachos, Baumgartner, Hunter, Carpenter, Tsai, Dai, Liu, Chen, Sun, Katrenko, Adriaans, Blaschke, Torres, Neves, Nakov, Divoli, Ma{\~{n}}a-L{\'o}pez, Mata, and Wilbur}]{dataset:BC2GM}
Larry Smith, Lorraine~K. Tanabe, Rie Johnson~nee Ando, Cheng-Ju Kuo, I-Fang Chung, Chun-Nan Hsu, Yu-Shi Lin, Roman Klinger, Christoph~M. Friedrich, Kuzman Ganchev, Manabu Torii, Hongfang Liu, Barry Haddow, Craig~A. Struble, Richard~J. Povinelli, Andreas Vlachos, William~A. Baumgartner, Lawrence Hunter, Bob Carpenter, Richard Tzong-Han Tsai, Hong-Jie Dai, Feng Liu, Yifei Chen, Chengjie Sun, Sophia Katrenko, Pieter Adriaans, Christian Blaschke, Rafael Torres, Mariana Neves, Preslav Nakov, Anna Divoli, Manuel Ma{\~{n}}a-L{\'o}pez, Jacinto Mata, and W.~John Wilbur. 2008.
\newblock \href {https://doi.org/10.1186/gb-2008-9-s2-s2} {Overview of biocreative ii gene mention recognition}.
\newblock \emph{Genome Biology}, 9(2):S2.

\bibitem[{Sun et~al.(2021)Sun, Yang, Wang, Zhang, Lin, and Wang}]{sun_MRC}
Cong Sun, Zhihao Yang, Lei Wang, Yin Zhang, Hongfei Lin, and Jian Wang. 2021.
\newblock \href {https://www.sciencedirect.com/science/article/pii/S1532046421001283} {Biomedical named entity recognition using bert in the machine reading comprehension framework}.
\newblock \emph{Journal of Biomedical Informatics}, 118:103799.

\bibitem[{Sänger et~al.(2024)Sänger, Garda, Wang, Weber-Genzel, Droop, Fuchs, Akbik, and Leser}]{HunFlair2}
Mario Sänger, Samuele Garda, Xing~David Wang, Leon Weber-Genzel, Pia Droop, Benedikt Fuchs, Alan Akbik, and Ulf Leser. 2024.
\newblock \href {https://doi.org/10.1093/bioinformatics/btae564} {Hunflair2 in a cross-corpus evaluation of biomedical named entity recognition and normalization tools}.
\newblock \emph{Bioinformatics}, 40(10):btae564.

\bibitem[{Tan et~al.(2021)Tan, Shen, Zhang, Lu, and Zhuang}]{ijcai2021p0542}
Zeqi Tan, Yongliang Shen, Shuai Zhang, Weiming Lu, and Yueting Zhuang. 2021.
\newblock \href {https://doi.org/10.24963/ijcai.2021/542} {A sequence-to-set network for nested named entity recognition}.
\newblock In \emph{Proceedings of the Thirtieth International Joint Conference on Artificial Intelligence, {IJCAI-21}}, pages 3936--3942. International Joint Conferences on Artificial Intelligence Organization.
\newblock Main Track.

\bibitem[{Tjong Kim~Sang and De~Meulder(2003)}]{conll_tjong-kim-sang-de-meulder-2003-introduction}
Erik~F. Tjong Kim~Sang and Fien De~Meulder. 2003.
\newblock \href {https://aclanthology.org/W03-0419/} {Introduction to the {C}o{NLL}-2003 shared task: Language-independent named entity recognition}.
\newblock In \emph{Proceedings of the Seventh Conference on Natural Language Learning at {HLT}-{NAACL} 2003}, pages 142--147.

\bibitem[{Wang et~al.(2021)Wang, Jiang, Bach, Wang, Huang, Huang, and Tu}]{ACE_wang-etal-2021-automated}
Xinyu Wang, Yong Jiang, Nguyen Bach, Tao Wang, Zhongqiang Huang, Fei Huang, and Kewei Tu. 2021.
\newblock \href {https://doi.org/10.18653/v1/2021.acl-long.206} {Automated concatenation of embeddings for structured prediction}.
\newblock In \emph{Proceedings of the 59th Annual Meeting of the Association for Computational Linguistics and the 11th International Joint Conference on Natural Language Processing (Volume 1: Long Papers)}, pages 2643--2660, Online. Association for Computational Linguistics.

\bibitem[{Wang et~al.(2018)Wang, Zhang, Ren, Zhang, Zitnik, Shang, Langlotz, and Han}]{cross_type_wang}
Xuan Wang, Yu~Zhang, Xiang Ren, Yuhao Zhang, Marinka Zitnik, Jingbo Shang, Curtis Langlotz, and Jiawei Han. 2018.
\newblock \href {https://doi.org/10.1093/bioinformatics/bty869} {Cross-type biomedical named entity recognition with deep multi-task learning}.
\newblock \emph{Bioinformatics}, 35(10):1745--1752.

\bibitem[{Wei et~al.(2022)Wei, Allot, Riehle, Milosavljevic, and Lu}]{dataset:TmVar3}
Chih-Hsuan Wei, Alexis Allot, Kevin Riehle, Aleksandar Milosavljevic, and Zhiyong Lu. 2022.
\newblock \href {https://doi.org/10.1093/bioinformatics/btac537} {tmvar 3.0: an improved variant concept recognition and normalization tool}.
\newblock \emph{Bioinformatics}, 38(18):4449--4451.

\bibitem[{Wei et~al.(2015)Wei, Kao, and Lu}]{dataset:GNormPlus}
Chih-Hsuan Wei, Hung-Yu Kao, and Zhiyong Lu. 2015.
\newblock {GNormPlus}: An integrative approach for tagging genes, gene families, and protein domains.
\newblock \emph{Biomed Res Int}, 2015:918710.

\bibitem[{Wolf et~al.(2020)Wolf, Debut, Sanh, Chaumond, Delangue, Moi, Cistac, Rault, Louf, Funtowicz, Davison, Shleifer, von Platen, Ma, Jernite, Plu, Xu, Le~Scao, Gugger, Drame, Lhoest, and Rush}]{wolf-etal-2020-transformers}
Thomas Wolf, Lysandre Debut, Victor Sanh, Julien Chaumond, Clement Delangue, Anthony Moi, Pierric Cistac, Tim Rault, Remi Louf, Morgan Funtowicz, Joe Davison, Sam Shleifer, Patrick von Platen, Clara Ma, Yacine Jernite, Julien Plu, Canwen Xu, Teven Le~Scao, Sylvain Gugger, Mariama Drame, Quentin Lhoest, and Alexander Rush. 2020.
\newblock \href {https://doi.org/10.18653/v1/2020.emnlp-demos.6} {Transformers: State-of-the-art natural language processing}.
\newblock In \emph{Proceedings of the 2020 Conference on Empirical Methods in Natural Language Processing: System Demonstrations}, pages 38--45, Online. Association for Computational Linguistics.

\bibitem[{Yan et~al.(2023)Yan, Sun, Li, and Qiu}]{yan-etal-2023-embarrassingly}
Hang Yan, Yu~Sun, Xiaonan Li, and Xipeng Qiu. 2023.
\newblock \href {https://doi.org/10.18653/v1/2023.acl-short.123} {An embarrassingly easy but strong baseline for nested named entity recognition}.
\newblock In \emph{Proceedings of the 61st Annual Meeting of the Association for Computational Linguistics (Volume 2: Short Papers)}, pages 1442--1452, Toronto, Canada. Association for Computational Linguistics.

\bibitem[{Yasunaga et~al.(2022)Yasunaga, Leskovec, and Liang}]{yasunaga-etal-2022-linkbert}
Michihiro Yasunaga, Jure Leskovec, and Percy Liang. 2022.
\newblock \href {https://doi.org/10.18653/v1/2022.acl-long.551} {{L}ink{BERT}: Pretraining language models with document links}.
\newblock In \emph{Proceedings of the 60th Annual Meeting of the Association for Computational Linguistics (Volume 1: Long Papers)}, pages 8003--8016, Dublin, Ireland. Association for Computational Linguistics.

\bibitem[{Yin et~al.(2024)Yin, Kim, Xiao, Wei, Kang, Lu, Xu, Fang, and Chen}]{YIN2024104731}
Yu~Yin, Hyunjae Kim, Xiao Xiao, Chih~Hsuan Wei, Jaewoo Kang, Zhiyong Lu, Hua Xu, Meng Fang, and Qingyu Chen. 2024.
\newblock \href {https://doi.org/10.1016/j.jbi.2024.104731} {Augmenting biomedical named entity recognition with general-domain resources}.
\newblock \emph{Journal of Biomedical Informatics}, 159:104731.

\bibitem[{Yu et~al.(2020)Yu, Bohnet, and Poesio}]{yu-etal-2020-named}
Juntao Yu, Bernd Bohnet, and Massimo Poesio. 2020.
\newblock \href {https://doi.org/10.18653/v1/2020.acl-main.577} {Named entity recognition as dependency parsing}.
\newblock In \emph{Proceedings of the 58th Annual Meeting of the Association for Computational Linguistics}, pages 6470--6476, Online. Association for Computational Linguistics.

\bibitem[{Zhang et~al.(2023)Zhang, Cheng, Gao, and Poon}]{zhang2023optimizing}
Sheng Zhang, Hao Cheng, Jianfeng Gao, and Hoifung Poon. 2023.
\newblock \href {https://openreview.net/forum?id=9EAQVEINuum} {Optimizing bi-encoder for named entity recognition via contrastive learning}.
\newblock In \emph{The Eleventh International Conference on Learning Representations}.

\end{thebibliography}
\clearpage
\appendix
\section{Datasets and Experimental Setup}
\label{appendix:datasets}

For the English subset of CoNLL-2003 \cite{conll_tjong-kim-sang-de-meulder-2003-introduction}, the original dataset splits are used, which are provided in a pre-tokenized format. For the GENIA dataset, the splits from \citealp{yan-etal-2023-embarrassingly} are adopted. The entity counts per split of these datasets can be found in Table \ref{tab:ST_entity_counts}.

\begin{table}[H]
    \centering
    \small
    \begin{tabular}{llrrr}
        \toprule
        \textbf{Dataset}       & \textbf{Entity Type} & \textbf{Train} & \textbf{Dev} & \textbf{Test} \\ \midrule
\multirow{4}{*}{CONLL} & LOC                  & 7,140          & 1,837        & 1,668         \\
                       & MISC                 & 3,438          & 922          & 702           \\
                       & ORG                  & 6,321          & 1,341        & 1,661         \\
                       & PER                  & 6,600          & 1,842        & 1,617         \\ \midrule
\multirow{5}{*}{GENIA} & Cell Line            & 3,069          & 372          & 403           \\
                       & Cell Type            & 5,854          & 576          & 578           \\
                       & DNA                  & 7,707          & 1,161        & 1,132         \\
                       & Gene or protein      & 28,874         & 2,466        & 2,900         \\
                       & RNA                  & 699            & 139          & 106           \\ \bottomrule
    \end{tabular}
    \caption{Statistics for the datasets used in the single-task experiments of section \ref{sec:ST_results}.}
    \label{tab:ST_entity_counts}
\end{table}

To train a multi-task model, six biomedical datasets are utilized: BC2GM \cite{dataset:BC2GM}, BC4CHEMD \cite{dataset:BC4CHEMD}, BC5CDR \cite{dataset:BC5CDR}, JNLPBA \cite{dataset:JNLPBA}, Linnaeus \cite{dataset:Linnaeus}, and NCBI Disease \cite{dataset:NCBIDisease}. The dataset splits (Table \ref{tab:MT_entity_counts}) follow those established by \citealp{crichton_neural_2017}, which have been extensively used in prior studies, including \citealp{cross_type_wang, huang-etal-2019-learning, khan2020mtbionermultitasklearningbiomedical, taughtnet}.

\begin{table}[H]
    \centering
    \scriptsize
    \begin{tabular}{llrrr}
        \toprule
        \textbf{Dataset} & \textbf{Entity Type} & \textbf{Train} & \textbf{Dev} & \textbf{Test} \\
        \midrule
        \multirow{1}{*}{BC2GM} & Gene or protein & 15,035 & 3,032 & 6,243 \\
        \midrule
        \multirow{1}{*}{BC4CHEMD} & Chemical & 29,263 & 29,305 & 25,210 \\
        \midrule
        \multirow{2}{*}{BC5CDR} & Chemical & 5,114 & 5,239 & 5,277 \\
        & Disease & 4,169 & 4,224 & 4,394 \\
        \midrule
        \multirow{5}{*}{JNLPBA} & Cell Line & 3,369 & 389 & 490 \\
        & Cell Type & 6,162 & 522 & 1,906 \\
        & DNA & 8,416 & 1,040 & 1,045 \\
        & Gene or protein & 27,015 & 2,379 & 4,988 \\
        & RNA & 844 & 106 & 118 \\
        \midrule
        \multirow{1}{*}{Linnaeus} & Species & 2,079 & 700 & 1,412 \\
        \midrule
        \multirow{1}{*}{NCBI Disease} & Disease & 5,111 & 779 & 952 \\
        \bottomrule
    \end{tabular}
    \caption{Statistics for the datasets used in the multi-task experiment of section \ref{sec:results_MT}.}
    \label{tab:MT_entity_counts}
\end{table}

In the aforementioned experiments, models are trained on the respective training splits, checkpoint selection is made on the development splits, and evaluation is conducted on the test splits.

For the cross-corpus evaluation, the experimental setup of \citealp{HunFlair2} is replicated. A multi-task model is trained using an ensemble of nine datasets\footnote{The datasets were obtained in February 2025 from \url{https://github.com/flairnlp/flair}. Their splits and preprocessing choices were replicated.}: BioRED \cite{dataset:BioRED}, GNormPlus \cite{dataset:GNormPlus}, Linnaeus \cite{dataset:Linnaeus}, NCBI Disease \cite{dataset:NCBIDisease}, NLM-Chem \cite{dataset:NLMchem}, NLM-Gene \cite{dataset:NLMGene}, S800 \cite{dataset:S800}, SCAI Chemical \cite{dataset:SCAIchemical}, and SCAI Disease \cite{dataset:SCAIdisease}. The model is trained on the training sets, with checkpoint selection being performed on the development splits. The evaluation is conducted on an independent corpus consisting of the full annotated data of three datasets\footnote{The preprocessed datasets were downloaded from \url{https://github.com/hu-ner/hunflair2-experiments} in February 2025.}: BioID \cite{dataset:BioID}, MedMentions \cite{dataset:MedMentions}, and tmVar3 \cite{dataset:TmVar3}. Dataset statistics for the training corpora and the independent test corpora can be found in Table \ref{tab:HF_train_entity_counts} and Table \ref{tab:HF_eval_entity_counts}, respectively.

\begin{table}[H]
    \centering
    \scriptsize
    \begin{tabular}{llrrr}
        \toprule
        \textbf{Dataset} & \textbf{Entity Type} & \textbf{Train} & \textbf{Dev} & \textbf{Test} \\
        \midrule
        \multirow{5}{*}{BioRED} & Cell Line & 103 & 22 & 50 \\
                                & Chemical & 2,830 & 818 & 751 \\
                                & Disease & 3,643 & 982 & 917 \\
                                & Gene & 4,404 & 1,087 & 1,170 \\
                                & Species & 1,429 & 370 & 393 \\
        \midrule
        \multirow{1}{*}{GNormPlus} & Gene & 4,964 & 504 & 4,468 \\
        \midrule
        \multirow{1}{*}{Linneaus} & Species & 1,725 & 206 & 793 \\
        \midrule
        \multirow{1}{*}{NCBI Disease} & Disease & 4,083 & 666 & 2,109 \\
        \midrule
        \multirow{1}{*}{NLM-Chem} & Chemical & 21,102 & 5,223 & 11,571 \\
        \midrule
        \multirow{1}{*}{NLM-Gene} & Gene & 11,209 & 1,314 & 2,687 \\
        \midrule
        \multirow{1}{*}{S800} & Species & 2,236 & 410 & 1,079 \\
        \midrule
        \multirow{1}{*}{SCAI Chemical} & Chemical & 852 & 83 & 375 \\
        \midrule
        \multirow{1}{*}{SCAI Disease} & Disease & 1,281 & 250 & 710 \\
        \bottomrule
    \end{tabular}
    \caption{Statistics of the training corpora used in the cross-corpus evaluation scenario of section \ref{sec:results_CC}.}
    \label{tab:HF_train_entity_counts}
\end{table}

\begin{table}[H]
    \centering
    \small
    \begin{tabular}{llr}
        \toprule
        \textbf{Dataset} & \textbf{Entity Type} & \textbf{Number of mentions} \\
        \midrule
        \multirow{1}{*}{BioID} & Species & 7,939 \\
        \midrule
        \multirow{1}{*}{tmVar3} & Gene & 4,059 \\
        \midrule
        \multirow{2}{*}{MedMentions} & Disease & 19,298 \\
        & Chemical & 19,198 \\
        \bottomrule
    \end{tabular}
    \caption{Statistics of the corpora used for the cross-corpus evaluation described in section \ref{sec:results_CC}.}
    \label{tab:HF_eval_entity_counts}
\end{table}

\par Finally, in order to assess the model's global prediction ability, synthetic datasets were derived from the BC5CDR corpus, in line with \cite{huang-etal-2019-learning} experimental setup. The original training set was randomly divided into two disjoint subsets: BC5-Disease (containing only Disease annotations) and BC5-Chemical (containing only Chemical annotations).  The same procedure was followed for the development splits. The statistics of these synthetic datasets can be found in Table \ref{tab:simulated_entity_counts}. By training models on the BC5-Disease and BC5-Chemical subsets and evaluating them on the full test split of the BC5CDR corpus, we can test the models global prediction abilities, as described in section \ref{sec:global_predictions}.

\begin{table}[H]
    \centering
    \small
    \begin{tabular}{llrr}
        \toprule
        \textbf{Dataset} & \textbf{Entity Type} & \textbf{Train} & \textbf{Dev} \\
        \midrule
        \multirow{1}{*}{BC5-Disease} & Disease & 2,172 & 2,279 \\
        \midrule
        \multirow{1}{*}{BC5-Chemical} & Chemical & 2,459 & 2,665 \\
        \bottomrule
    \end{tabular}
    \caption{Statistics of the synthetic datasets created for assessing global prediction ability.}
    \label{tab:simulated_entity_counts}
\end{table}

\section{Training Details}
\label{appendix:training}

\begin{table}[H]
    \centering
    \small
    \begin{tabular}{lrr}
    \toprule
\textbf{Hyperparameter}                   & \textbf{GENIA} & \textbf{Others} \\ \midrule
\# epochs                        & 100 & 100   \\
Early stop                       & 30 & 30    \\
Batch size                       & 16 & 16   \\
Max. \# tokens                    & 405 & 405  \\
Gradient norm clipping           & 1.0 & 1.0  \\
Dropout on logits                & 0.1 & 0.1  \\ \midrule
\textit{SRU module}              &       \\ \midrule
\# latent embeddings (multiplier) & 10 & 2 \\
Half-context for pos. embeddings    & 240 & 150  \\
Dropout on pos. embeddings          & 0.2 & 0.2  \\
Dropout on latent embeddings        & 0.2 & 0.2  \\ \midrule
\textit{Encoder optimizer}       &       \\ \midrule
LR                               & 3e-5 & 2e-5 \\
Weight decay                     & 1e-3 & 1e-3 \\
Warm up (in epochs)              & 1  & 1 \\ \midrule
\textit{Actions generation cycle optimizer} &       \\ \midrule
LR                               & 3e-4 & 3e-4 \\
Weight decay                     & 1e-3 & 1e-3 \\
Warm up (in epochs)              & 0.5  & 0.5   \\ \bottomrule
    \end{tabular}
    \caption{Hyperparameters used for the experiments. The column 'Others' refers to every experiment except the single-task on the GENIA dataset.}
    \label{tab:hyper}
\end{table}

All models are developed using the PyTorch tensor library and trained on a single NVIDIA A100 80GB GPU. The encoder module and the action generation module are tuned using two separate AdamW optimizers with linear warm-up, set with different initial learning rates and weight decays. Both optimizers are set with $\beta_1=0.9$, $\beta_2=0.98$ and $\epsilon=10^{-6}$. Models are trained with early stopping based on performance on the development set.\footnote{In the case of multi-task models where multiple datasets are tagged with the same entity type (the models of sections \ref{sec:results_MT} and \ref{sec:results_CC}), despite the entity types being considered disjoint for training purposes, validation F1 scores on the development set for checkpoint selection are computed by merging the types, as described in the begining of section \ref{sec:results_MT}.} The hyperparameters of all experiments can be found in Table \ref{tab:hyper}. Additionally, while the token scaling parameter $\alpha$ in equation (\ref{eq:sum_pos_embeds}) of section \ref{sec:SRU} was trained for the single-task experiment on the GENIA dataset, it was frozen and set to $1$ for all other experiments.

\par The encoder module was built on top of the HuggingFace \textit{transformers} library \cite{wolf-etal-2020-transformers}. Specifically, the \texttt{LinkBERT-large} encoder from \citealp{yasunaga-etal-2022-linkbert} was used for all models trained with biomedical corpora, while the \texttt{xlm-roberta-large} encoder introduced by \citealp{conneau-etal-2020-unsupervised} was used for the single task model trained on the CoNLL-2003 dataset.

\section{Single-task performance on the datasets used for the cross-corpus experiment}
\label{appendix:CC}

\begin{table}[H]
    \centering
    \small
    \begin{tabular}{lrr}
    \toprule
    \textbf{Dataset}       & \textbf{\textit{Merged}} & \textbf{\textit{Disjoint}} \\ \midrule
BioRED           & 90.73                    & 90.90                      \\
GNormPlus        & 85.00                    & 86.00                      \\
Linnaeus         & 78.16                    & 92.23                      \\
NCBI Disease     & 85.69                    & 85.70                      \\
NLM-Chem         & 84.42                    & 85.65                      \\
NLM-Gene         & 88.35                    & 88.13                      \\
S800             & 74.24                    & 75.79                      \\
SCAI Chemical    & 85.21                    & 85.64                      \\
SCAI Disease     & 80.78                    & 82.14                      \\           \bottomrule
    \end{tabular}
    \caption{In-corpus micro-F1 scores for the model used in the cross-corpus evaluation experiment of section \ref{sec:results_CC}.}
    \label{tab:cross_train_results}
\end{table}

\section{Human evaluation of global predictions in a cross-domain setting}
\label{appendix:human_eval}

To assess the model’s ability to generalize across domains, three models were trained:
\begin{itemize}
    \item \textit{SRU-NER-CoNLL}: a single-task model trained on the CoNLL corpus;
    \item \textit{SRU-NER-BC5}: a single-task model trained on the BC5CDR corpus;
    \item \textit{SRU-NER-MTL}: a multi-task model trained on both corpora.
\end{itemize}

\noindent All models were trained using the \texttt{LinkBERT-large} encoder from \citealp{yasunaga-etal-2022-linkbert}. To evaluate cross-domain generalization, the models capable of recognizing general-domain entity types (\textit{SRU-NER-CoNLL} and \textit{SRU-NER-MTL}) were used to annotate the test split of the biomedical corpus, while the models trained on biomedical entity types (\textit{SRU-NER-BC5} and \textit{SRU-NER-MTL}) were used to annotate the test split of the general-domain corpus. Since gold annotations for these out-of-domain predictions were not available, two linguists manually assessed their correctness. Inter-annotator agreement per entity type is reported in Table \ref{tab:agreement}.

\begin{table}[h]
\small
\centering
\begin{tabular}{lr}
\toprule
\textbf{Entity}   & \textbf{Agreement} (\%) \\ \midrule
Chemical & 92.98          \\
Disease  & 91.09          \\
LOC      & 100.00         \\
ORG      & 87.76          \\
PER      & 88.89          \\ \bottomrule
\end{tabular}
\caption{Inter-annotator agreement for the evaluated entity types.}
\label{tab:agreement}
\end{table}

\par Based on the assessment of correct predicted spans by the two human annotators, a precision score was computed by taking the ratio of correctly identified spans to the total number of predicted spans, for each model, entity type and linguist. A simulated recall score per model, entity type and linguist was also computed by considering the total number of spans of each entity type that were considered correct by at least one of the annotators, across all the predictions made by the three models. Finally, precision and simulated recall scores per model and entity type were obtained by averaging across the two human annotators.
\par The results can be found in Table \ref{tab:human_eval}, in the main text. One can see that the precision scores of the multi-task model are higher than the single-task ones across all entity types, while the recall values of the multi-task model are worse for all entity types except ORG.

\par For reference, the in-corpus performance of the three models is present in Table \ref{tab:human_eval_in_corpus}.

\begin{table}[H]
\small
\centering
\begin{tabular}{lrr}
\toprule
\textbf{Model}         & \textbf{CoNLL} & \textbf{BC5CDR} \\ \midrule
\textit{SRU-NER-CoNLL} & 90.51 & ---    \\
\textit{SRU-NER-BC5}   & ---   & 90.61  \\
\textit{SRU-NER-MT}    & 91.01 & 90.51  \\ \bottomrule
\end{tabular}
\caption{In-corpus performance of the three models used for evaluation of global predictions in a cross-domain setting. The single-task model SRU-NER-BC5 is the same as the one used for comparison in the multi-task experiment of section \ref{sec:results_MT}.}
\label{tab:human_eval_in_corpus}
\end{table}

\end{document}